\documentclass[runningheads]{llncs}
\usepackage[utf8]{inputenc}    
\usepackage[T1]{fontenc}
\usepackage{enumitem}
\usepackage{todonotes}
\usepackage{wrapfig}
\usepackage{float}
\usepackage{placeins}
\usepackage{adjustbox}
\usepackage{calc}
\usepackage{amsmath}
\usepackage{listings}          
\usepackage{graphicx}
\lstdefinestyle{jsonstyle}{
    basicstyle=\ttfamily\small,
    breaklines=true,
    columns=fullflexible,
    showstringspaces=false,
    frame=single,
    literate=
      {∞}{{\ensuremath{\infty}}}1
      {→}{{\ensuremath{\to}}}1
      {≥}{{\ensuremath{\ge}}}1
      {∪}{{\ensuremath{\cup}}}1
}
\begin{document}
\title{From Struggle (06-2024) to Mastery (02-2025) LLMs Conquer Advanced Algorithm Exams and Pave the Way for Editorial Generation}
%
%
\author{Adrian Marius Dumitran\inst{1}\inst{3}\orcidID{0009-0005-3547-5772} \and \\
Theodor-Pierre Moroianu\inst{2}\orcidID{0009-0001-5041-2201} \and
Vasile Paul Alexe\inst{1}\orcidID{0009-0000-9442-4944}}
\authorrunning{A. Dumitran et al.}
%
\institute{University of Bucharest, Bucharest, Romania \\
\email{marius.dumitran@unibuc.ro, vasile-paul.alexe@g.unibuc.ro}
\and
ETH, Zurich, Switzerland \\
\email{theodor.moroianu@gmail.com}
\and
Softbinator Technologies, Bucharest, Romania}
\maketitle              
\begin{abstract}
This paper presents a comprehensive evaluation of the performance of state-of-the-art Large Language Models (LLMs) on challenging university-level algorithms exams. By testing multiple models on both a Romanian exam and its high-quality English translation, we analyze LLMs' problem-solving capabilities, consistency, and multilingual performance. Our empirical study reveals that the most recent models not only achieve scores comparable to top-performing students but also demonstrate robust reasoning skills on complex, multi-step algorithmic challenges, even though difficulties remain with graph-based tasks. Building on these findings, we explore the potential of LLMs to support educational environments through the generation of high-quality editorial content, offering instructors a powerful tool to enhance student feedback. The insights and best practices discussed herein pave the way for further integration of generative AI in advanced algorithm education.


\end{abstract}
\section{Introduction}

Recent breakthroughs in large language models (LLMs) have dramatically expanded their capabilities across various domains. Historically, although LLMs excelled in many areas, their performance on advanced algorithm problems was limited. Over the past year, however, significant improvements have been made, enabling newer models to solve complex algorithmic challenges with remarkable accuracy. This is highlighted very clearly in this 2025 paper about OpenAi's o3 model, which reaches extraordinary levels in competitive programming \cite{ref_openai2025competitive}.

In this paper, we conduct an in-depth analysis of these advancements within the context of a university-level advanced algorithms exam. Our empirical results reveal that models older than six months tend to perform in the bottom 40\% of the ranking, whereas recent models from January–February 2025 consistently rank within the top 15\%, with the new o3-mini even achieving top 5\% status. We show these improvements in Section \ref{sec:llm_solvers}, where we focus on evaluating LLM performance as exam solvers in real-world scenarios. We also highlight areas where LLMs still underperform compared to average students.

Additionally, we explore two promising applications of LLM technology in educational settings. First, we present a human-in-the-loop approach for generating detailed grading schemes, enabling instructors to efficiently grade student work. Second, we examine how LLMs can assist in generating high-quality editorial content, providing clear and actionable feedback to students. We also offer a free-to-use framework for this task.

\section{Related Work}
\label{sec:related_work}

Our work builds upon research at the intersection of Large Language Models (LLMs), algorithmic problem-solving, and educational applications.  LLMs are increasingly being explored for their ability to tackle complex tasks, as evidenced by their application in competitive programming environments \cite{ref_openai2025competitive}.  

The use of LLMs in education is a rapidly developing area, with studies investigating their potential for grading and assessment \cite{Lee2024a,Lee2024b} and automatic feedback generation \cite{Furuhashi2025}. Recent efforts, like that of \cite{Xie2024}, also focus on improving automated grading systems with LLMs, addressing issues such as rubric generation, scoring consistency, and introducing "Grade-Like-a-Human" systems that mimic human evaluation through multi-agent approaches. The impact of LLMs on education is also being broadly considered, examining both opportunities and challenges for various stakeholders \cite{Kasneci2023}.

This work differs from existing research by focusing on advanced STEM content in a real-world exam setting, evaluating grading consistency and performance on the content on Romanian algorithmic exams and by providing a novel dataset in the low-resource Romanian language, with a similar study being done by Anton, A.~et al.~\cite{Anton2024} for Brazilian exams and by Dumitran, A.~M.~et al.~\cite{Dumitran2024} on Romanian competitive programming.

\section{LLMs as Exam Solvers: An Empirical Evaluation}

\label{sec:llm_solvers}

We start by conducting a thorough evaluation of their problem-solving capabilities in the context of a challenging computer science exam, composed of 11 algorithmic problems. This exam, which we release as a publicly available dataset \cite{ref_github}, covers advanced algorithms with a focus on graphs  and was originally administered in Romanian. This chapter details our methodology and presents the results of this evaluation, focusing on the performance of a diverse range of state-of-the-art LLMs.

\subsection{Methodology}
\label{sec:methodology}

We evaluate a broad range of LLMs to assess their exam-solving abilities. Our selection includes models with diverse architectures and training methods. We prioritize those accessible via user interfaces—mimicking a typical instructor’s approach—while also including deprecated models to illustrate that older models are ineffective as tutoring partners.

Models were selected from the ChatBot Arena LLM Leaderboard~\cite{ref_chatbot}, choosing a few from each family. From the OpenAI UI \cite{chatgpt}, we evaluated \texttt{GPT-4 Legacy}, \texttt{GPT-4o}, \texttt{o1}, \texttt{o3-mini} (Jan 31, 2025), and \texttt{o3-mini-high} (Jan 31, 2025). From Google AI Studio \cite{google-ai-studio}, we assessed \texttt{Gemini 2.0 Flash} (Feb 2025), \texttt{Gemini 1.5 Pro}, \texttt{Gemini 2.0 Pro Experimental} (Feb 2025), and \texttt{Gemini 2.0 Flash Thinking Experimental} (Jan 2025), as well as the open-source \texttt{Gemma 2-27B}. Using the \textit{Together AI} platform \cite{together-ai}, we also evaluated \texttt{DeepSeekR1}, \texttt{Qwen2.5Max} (Jan 2025), \texttt{Claude Sonnet 3.5}, and models from Meta, namely \texttt{Llama3.3-70B}, \texttt{Llama-3.1-405B}, \texttt{Mistral 7B Inst-v0.3}, and \texttt{Mixtral 22x8 Inst-v0.1}. In this paper, we will refer to each LLM by its model name without including its brand. Thus, ‘OpenAi’s o3-mini’ will be referred to simply as ‘o3-mini’, and ‘Claude Sonnet 3.5’ as ‘Sonnet 3.5’..

For most models, we use their official user interfaces with default settings and a simple, neutral prompt, reflecting an “out-of-the-box” usage scenario.

The exam, originally in Romanian, is also translated into high-quality English. Crucially, the LLMs are instructed to solve the entire exam in a single interaction, i.e. one-shot, rather than addressing each of the 11 problems individually. This design choice is made to evaluate the models’ ability to manage long contexts and maintain coherence across a complex, multi-part task -- a key requirement for real-world exam solving. All LLM responses are scored by the course instructor, ensuring consistency and fairness with the grades received by the students who took the exam.

\subsection{New Models Triumph Where Old Ones Fail}

\begin{figure}[htbp]
    \centering
    \boxed{\includegraphics[width=0.8\textwidth]{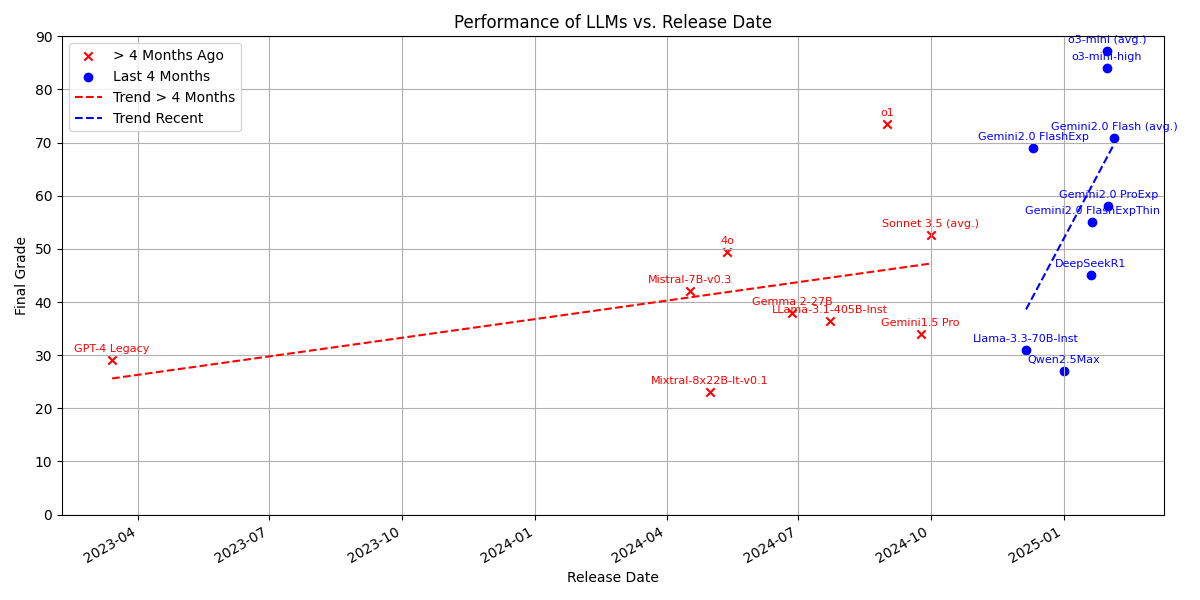}}
    \caption{LLM Performance on Advanced Algorithms Exam vs. Release Date. Models are grouped by release date: more than four months ago (red line, crosses) and within the last four months (blue line, circles).}
    \label{fig:llm_performance}
\end{figure}

Leading-edge LLMs' performance on the advanced algorithms exam is strongly correlated with the models' release dates. As shown in Figure~\ref{fig:llm_performance}, models released within the last four months such as OpenAI's \texttt{o3-mini} and Google's \texttt{Gemini2.0 Flash} consistently score above 80 and 70 points (top 5\% and 15\% respectively) on our evaluation exam. In contrast, older models often score below 40. 

The older models struggle not only with intricate algorithmic problems but also with processing the extensive context of the exam. This suggests that handling larger and more complex context windows is a key differentiator between newer and older LLMs. Although isolated exceptions exist, the overall trend strongly indicates that recent LLM development has crossed a critical threshold, making them effective tools for challenging educational assessments in advanced computer science topics. Consequently, this period represents a crucial juncture for investigating the potential of LLMs in advanced algorithm courses, as recent breakthroughs have dramatically enhanced their applicability.

It is also important to mention that neither \texttt{DeepSeekR1} nor the new \texttt{Qwen 2.5 Pro} perform well; both fail the exam and rank in the bottom half. Additionally, high server load hindered \texttt{DeepSeekR1}'s usability via both the UI and API.

\subsection{Consistency Analysis of LLMs in Exam Grading}

As our goal is to assess LLMs for tutoring we examine their consistency in solving the exam's problems. For this experiment we select the models obtaining good results in our initial study: we are not interested in using as a tutor an LLM that is consistently wrong.
This subsection examines the consistency of LLMs in grading exams by evaluating three representative models: \texttt{o3-mini}, \texttt{Gemini 2.0 Flash}, and \texttt{Sonnet 3.5}. Consistency is measured by the standard deviation (SD) of their final grades across multiple runs -- a lower SD indicates higher consistency.

Our methodology for measuring the consistency of the 3 models is the following:

\begin{itemize}
    \item We ask each model to solve the exam in 5 different sessions.
    \item We grade each solution independently.
    \item For each model we compute the mean and SD of its 5 solutions.
\end{itemize}

The consistency results can be seen in the Table \ref{tab:consistency}.

\begin{table}[h]
\centering
\begin{tabular}{|l|c|c|c|}
\hline
\textbf{Model} & \textbf{Runs} & \textbf{Final Grade Mean} & \textbf{Final Grade SD} \\
\hline
o3-mini & 5 & 86.60 & 2.41 \\
Gemini 2.0 Flash & 5 & 69.90 & 2.70 \\
Sonnet 3.5 & 5 & 53.70 & 16.95 \\ 
\hline
\end{tabular}
\vspace{10px}
\caption{Final grade consistency across multiple runs.}
\label{tab:consistency}
\end{table}

We observe that \texttt{o3-mini} and \texttt{Gemini 2.0 Flash} exhibit significantly higher consistency, as evidenced by their low standard deviations (2.41 and 2.70, respectively), compared to Sonnet 3.5, which shows substantial variability with an SD of 16.95. A further analysis reveals that the scores of \texttt{Sonnet 3.5} are quite similar in 4 of the runs at around 50 points, but one of the runs scores an excellent 77 points, like a black swan situation. 

\subsection{Collaborative Solving}
\label{subsec:collaborative}

LLMs are trained on different datasets and thus show varying strengths. While the \texttt{o3-mini} model generally performs well, the \texttt{Gemini} family of models excels in theoretical tasks and competitive programming. We explore whether combining multiple solutions could yield a more effective solver.

We run two experiments, in which models are tasked with solving the exam:
\begin{itemize}
    \item \textbf{RunAvg}: Each LLM is provided with its own previous answers (from 3 different runs).
    \item \textbf{RunAvgAll}: Each LLM is provided with three solutions from different LLMs (including \texttt{DeepSeekR1} to ensure diversity).
\end{itemize}

 Table~\ref{tab:runavg} shows the difference between the scores obtained using these strategies and the mean individual scores.

\begin{table}[h]
\centering
\begin{tabular}{|l|c|c|}
\hline
\textbf{Model} & \textbf{RunAvg (Total)} & \textbf{RunAvgAll (Total)} \\
\hline
o3-mini      & 0.67  & -0.33 \\
Gemini 2.0 Flash    & -2.40 & 12.10 \\
Sonnet 3.5   & -8.67 & 13.83 \\
DeepSeekR1          & 0.50  & 15.00 \\
\hline
\end{tabular}
\vspace{10px}
\caption{Impact of RunAvg and RunAvgAll prompting strategies on scores.}
\label{tab:runavg}
\end{table}
\vspace{-20px}

The \emph{RunAvg} strategy shows little change, as self-reinforcement tends to maintain standard behavior. The only substantial difference is observed for \texttt{Sonnet 3.5}, where the score converges towards its mean, effectively reducing the impact of its outlier-level performance. 

In contrast, the \emph{RunAvgAll} strategy resulted in a substantial increase for \texttt{Gemini 2.0 Flash}, \texttt{Sonnet 3.5}, and \texttt{DeepSeekR1}, and a minor decrease for \texttt{o3-mini}. This outcome is understandable since the first models benefit from exposure to the better solutions of \texttt{o3-mini}, whereas \texttt{o3-mini} only gained access to lower-quality results.

Notably, \texttt{o3-mini} is able to leverage \texttt{Gemini 2.0 Flash}'s solution to optimally solve one of the problems (by employing a \textit{BFS 0-1} algorithm instead of Dijkstra's), although its max-flow solution suffered, preventing overall improvement.

\subsection{Comparison of LLM Scores and Student Performance}

\begin{figure}[h]
\centering
\boxed{\includegraphics[width=0.8\textwidth]{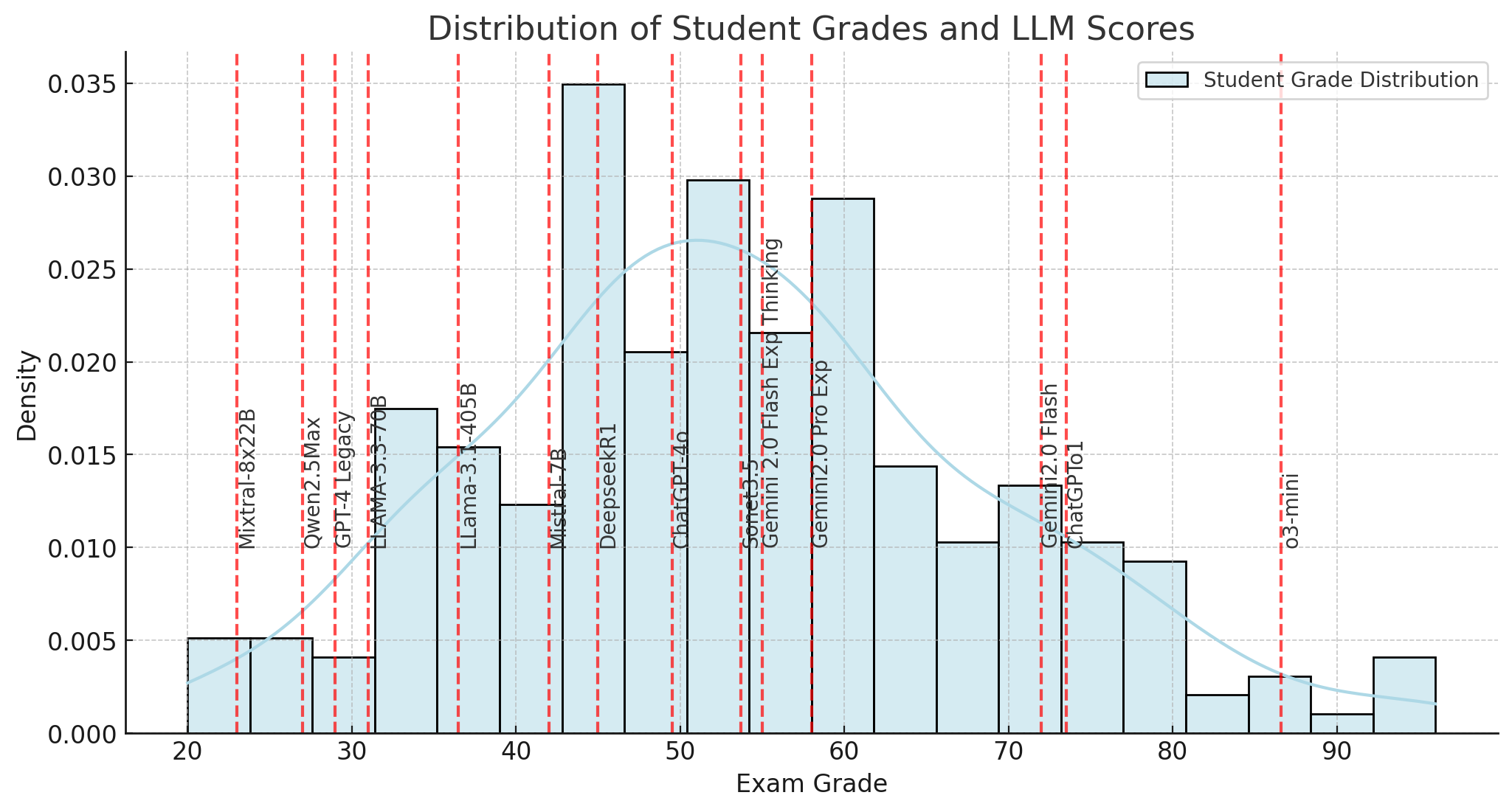}}
\caption{Distribution of student grades with LLM scores highlighted.}
\label{fig:grade_distribution}
\end{figure}

In this section, we compare the performance of various Large Language Models (LLMs) with student exam results. The distribution of student grades is presented in Figure~\ref{fig:grade_distribution}, where we also highlight the scores of different LLMs for a direct comparison.

From the figure, we observe that \texttt{o3-mini} and \texttt{Gemini2.0 Flash} perform well, achieving scores that place them among the top percentiles of student performance. In contrast, models such as \texttt{Mixtral-8x22B} and \texttt{GPT-4 Legacy} rank lower.

\begin{table}[h]
\centering
\footnotesize
\begin{tabular}{|l|c|c||l|c|c|}
\hline
\multicolumn{3}{|c||}{Top Half} & \multicolumn{3}{c|}{Bottom Half} \\
\hline
Model                        & Grade & Percentile & Model                         & Grade & Percentile \\
\hline
o3-mini                      & 86.60            & 98         & 4o                     & 49.50            & 40         \\
\hline
o3-mini-high                 & 84.00            & 97         & DeepSeekR1                    & 45.00            & 28         \\
\hline
o1                           & 73.50            & 89         & Gemma 2-27B                    & 38.00            & 17         \\
\hline
Gemini2.0Flash              & 72.30            & 88         & LLama3.1-405B-Inst       & 36.50            & 15         \\
\hline
Gemini2.0FlashExp & 69.00            & 83         & GPT-4 Legacy                    & 29.00            & 4          \\
\hline
Gemini2.0ProExp   & 58.00            & 64         & Qwen2.5Max                    & 27.00            & 4          \\
\hline
Sonnet 3.5                     & 53.70            & 53         & Mixtral-8x22B-Inst       & 23.00            & 2          \\
\hline
\end{tabular}
\caption{LLM Average Grades and Percentiles (Top Models on Left, Weaker on Right)}
\label{tab:llm_scores}
\end{table}

Table~\ref{tab:llm_scores} shows each LLM's standing within the student grade distribution, sorted by percentile ranking.

It is worth mentioning that around 5\% of the student cohort has previously participated in the National Olympiad in Computer Science, and the results obtained by \textit{o3-mini} places it not only top 3\% of the class but also, likely, top 1\% nationwide in computer science.

\subsection{LLMs VS Students Task Analysis}

We compare the results obtained by LLMs and students for each problem. The exam subjects are found in Appendix A.
\begin{table}[h]
\centering
\label{tab:student_llm_comparison}
\begin{tabular}{|c|c|c|c|}
\hline
Problem & Student Avg & LLM Avg & Difference \\
\hline
Problem 1 & 4.11 & 1.21 & \textbf{2.91} \\
Problem 2 & 4.21 & 3.59 & 0.62 \\
Problem 3 & 6.06 & 3.38 & \textbf{2.68} \\
Problem 4 & 4.30 & 3.21 & 1.09 \\
Problem 5 & 2.43 & 2.59 & -0.16 \\
Problem 6 & 4.34 & 2.76 & 1.58 \\
Problem 7 & 1.67 & 2.94 & -1.28 \\
Problem 8 & 7.68 & 2.82 & \textbf{4.85} \\
Problem 9 & 4.78 & 7.53 & -2.75 \\
Problem 10 & 1.96 & 4.32 & -2.36 \\
Problem 11 & 5.04 & 8.29 & -3.26 \\
\hline
\end{tabular}
\vspace{10px}
\caption{Comparison of Student and LLM Performance}
\end{table}

It is very interesting to note that the problems where LLMs perform significantly worse (such as Problems 1 and 3) are those that require visual analysis of a graph rather than the application of a typical algorithm.

\begin{figure}[h]
    \centering
    \boxed{\includegraphics[width=0.6\textwidth]{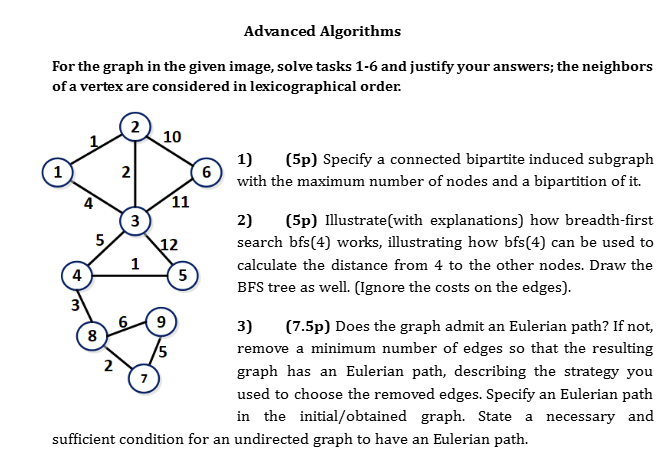}}
    \caption{Examples of visual tasks from Problems 1 and 3 that challenge LLMs.}
    \label{fig:problems1_3}
\end{figure}

In Figure~\ref{fig:problems1_3}, two such tasks are illustrated. While human students can easily solve visual tasks like these -- where one needs to, for example, visualize a bipartite graph or analyze the structure of a graph, LLMs face notable challenges. 

For instance, in Problem 1 most LLMs fail to correctly identify a valid solution. Below are a few sample outputs from various LLMs for this task:
\begin{itemize}
    \item \texttt{4o} outputs a long text, followed by: 
    \begin{quote}
    ``It seems that no sufficiently large bipartite component exists in the graph. The structure of the graph includes odd-length cycles that prevent direct bipartiteness. If you want, I can modify the graph by removing minimal edges to make it bipartite.''
    \end{quote}
    \item \texttt{DeepSeekR1} outputs an invalid solution:
    \begin{quote}
    ``A = \{4, 9\}, B = \{5, 8, 7\}.'' 
    \end{quote}
    As one can see from the drawing, vertices 8 and 7 are neighbors; thus, the solution is not only non-maximal but also invalid.
    \item \texttt{Gemini2.0 Flash} (one of the best models): 
    \begin{itemize}
        \item \textit{Set 1 (Color 0): \{1, 4, 7, 9\}}
        \item \textit{Set 2 (Color 1): \{2, 3, 8\}}
    \end{itemize}
    Again, this solution is not valid.
    \item \texttt{o3-mini} offers a valid solution:   
        \begin{itemize}
            \item \textit{Red:} \( \{3,8\} \)
            \item \textit{Blue:} \( \{1,4,6,7\} \)
        \end{itemize}
\end{itemize}

Problem 3 poses a similar challenge by asking for the minimal number of edges that must be removed from a given graph to allow for an Eulerian path. In this case, some of the newer thinking models—such as \texttt{DeepSeek} and \texttt{Gemini2.0 Flash Experimental Thinking} attempt backtracking but ultimately fail, often stalling after reaching the maximum number of output tokens.

Additionally, applying max-flow algorithms to a graph also seems to pose a challenge for LLMs (problem 8).

On the other hand, LLMs demonstrate strong performance in proving theoretical exercises (problems 7 and 9), applying string algorithms to concrete examples (exercise 10), and problem-solving (problem 11).

\subsection{Romanian vs.\ English Performance}

We evaluate each model on the same advanced algorithms exam in both Romanian and its English translation. We define the performance difference as the following:
\[
\Delta = \text{Grade}_{\text{En}} - \text{Grade}_{\text{Ro}},
\]
Note that a positive $\Delta$ indicates a higher score on the English version.

Table~\ref{tab:grouped_performance} shows each model’s release date, grades for the Romanian and English versions of the exam, and their delta. Models are sorted by release date and grouped (A vs.\ B) to highlight trends in more recent versus older systems. For models tested multiple times we present the average. Overall, newer models (Group~A) exhibit a small statistically unimportant difference in favor of Romanian ($-0.69$), while older models (Group~B) see larger gains in English ($+9.0$).

\begin{table}[h]
\centering
\small
\caption{Performance Comparison with Grouping (First 12 vs.\ Last 6)}
\label{tab:grouped_performance}
\begin{tabular}{|l|c|c|c|c|c|}
\hline
\textbf{Model}                 & \textbf{Release Date} & \textbf{Grade (Ro)} & \textbf{Grade (En)} & \textbf{Diff.} & \textbf{Group} \\
\hline
Gemini2.0 ProExp                & 2025-02    & 72.5  & 58.0  & -14.5 & A \\
Gemini2.0 Flash (avg.)          & 2025-02-05 & 61.0  & 70.8  & +9.8  & A \\
o3-mini (avg.)           & 2025-01-31 & 81.7  & 87.3  & +5.7  & A \\
o3-mini-high            & 2025-01-31 & 83.0  & 84.0  & +1.0  & A \\
Qwen2.5Max                     & 2025-01    & 34.0  & 27.0  & -7.0  & A \\
Gemini2.0 FlashExpThin         & 2025-01-21 & 61.0  & 55.0  & -6.0  & A \\
DeepSeekR1                     & 2025-01-20 & 46.5  & 45.0  & -1.5  & A \\
Gemini2.0 FlashExp              & 2024-12-11 & 68.5  & 69.0  & +0.5  & A \\
Llama-3.3-70B-Inst         & 2024-12-06 & 30.0  & 31.0  & +1.0  & A \\
Sonnet 3.5 (avg.)                & 2024-10    & 50.0  & 52.7  & +2.7  & A \\
Gemini1.5 Pro                   & 2024-09-24 & 40.0  & 34.0  & -6.0  & A \\
o1                      & 2024-09    & 67.5  & 73.5  & +6.0  & A \\
\hline
\multicolumn{1}{|c|}{\textbf{Group A Averages}} & 
\multicolumn{1}{c|}{After 09-2024} &
\textbf{58.0} &
\textbf{57.3} &
\multicolumn{2}{c|}{\textbf{-0.69}} \\
\hline
LLama-3.1-405B-Inst& 2024-07-23 & 34.0  & 36.5  & +2.5  & B \\
Gemma 2-27B                     & 2024-06-27 & 27.0  & 38.0  & +11.0 & B \\
4o                      & 2024-05-13 & 22.0  & 49.5  & +27.5 & B \\
Mixtral-8x22B-It-v0.1          & 2024-05    & 27.0  & 23.0  & -4.0  & B \\
Mistral-7B-v0.3                & 2024-04-17 & 23.0  & 42.0  & +19.0 & B \\
GPT-4 Legacy                     & 2023-03-14 & 31.0  & 29.0  & -2.0  & B \\
\hline
\multicolumn{1}{|c|}{\textbf{Group B Averages}} & 
\multicolumn{1}{c|}{Before 09-2024} &
\textbf{27.3} &
\textbf{36.3} &
\multicolumn{2}{c|}{\textbf{+9.0}} \\
\hline
\end{tabular}
\end{table}

Notably, older models like \texttt{4o} and \texttt{Mistral-7B-v0.3} show a strong bias toward English, reflecting potentially unbalanced multilingual data. Meanwhile, recent models such as \texttt{o3-mini}, \texttt{o3-mini-high} 
and \texttt{Gemini2.0 FlashExp}  have much smaller gaps, pointing to improved multilingual training.

\section{LLM-Assisted grading schemes and Editorial Generation}

\subsection{Grading Scheme Via Human-AI Collaboration}

We develop a grading scheme through a Human-AI collaboration between the course instructors and the top-performing LLM solvers, \texttt{o3-mini} and \texttt{Gemini Flash 2.0}. The process involves providing the LLMs with brief descriptions of solutions and prompting them to expand these into a comprehensive grading scheme, with instructors refining and correcting any errors along the way.

The result is an exceptionally detailed grading scheme, which can be found within the dataset on GitHub \cite{githubDataset}. The grading scheme is far more extensive than a typical one, as instructors would rarely have the time or energy to create such a thorough evaluation framework manually.

\subsection{Web-based Platform}

\begin{figure}
    \centering
    \includegraphics[width=0.6\textwidth]{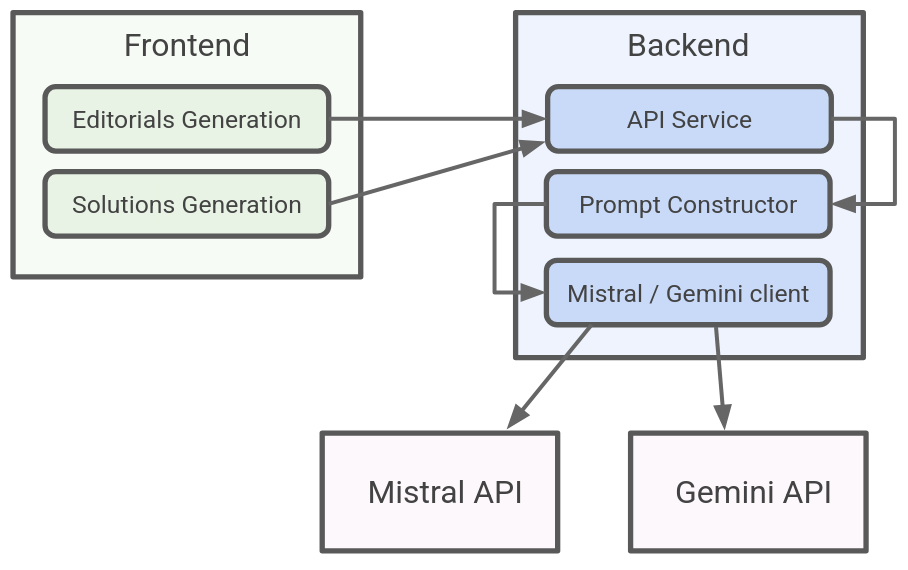}
    \caption{Design of our application.}
    \label{fig:design}
\end{figure}

\begin{figure}
    \centering
    \boxed{\includegraphics[width=\textwidth]{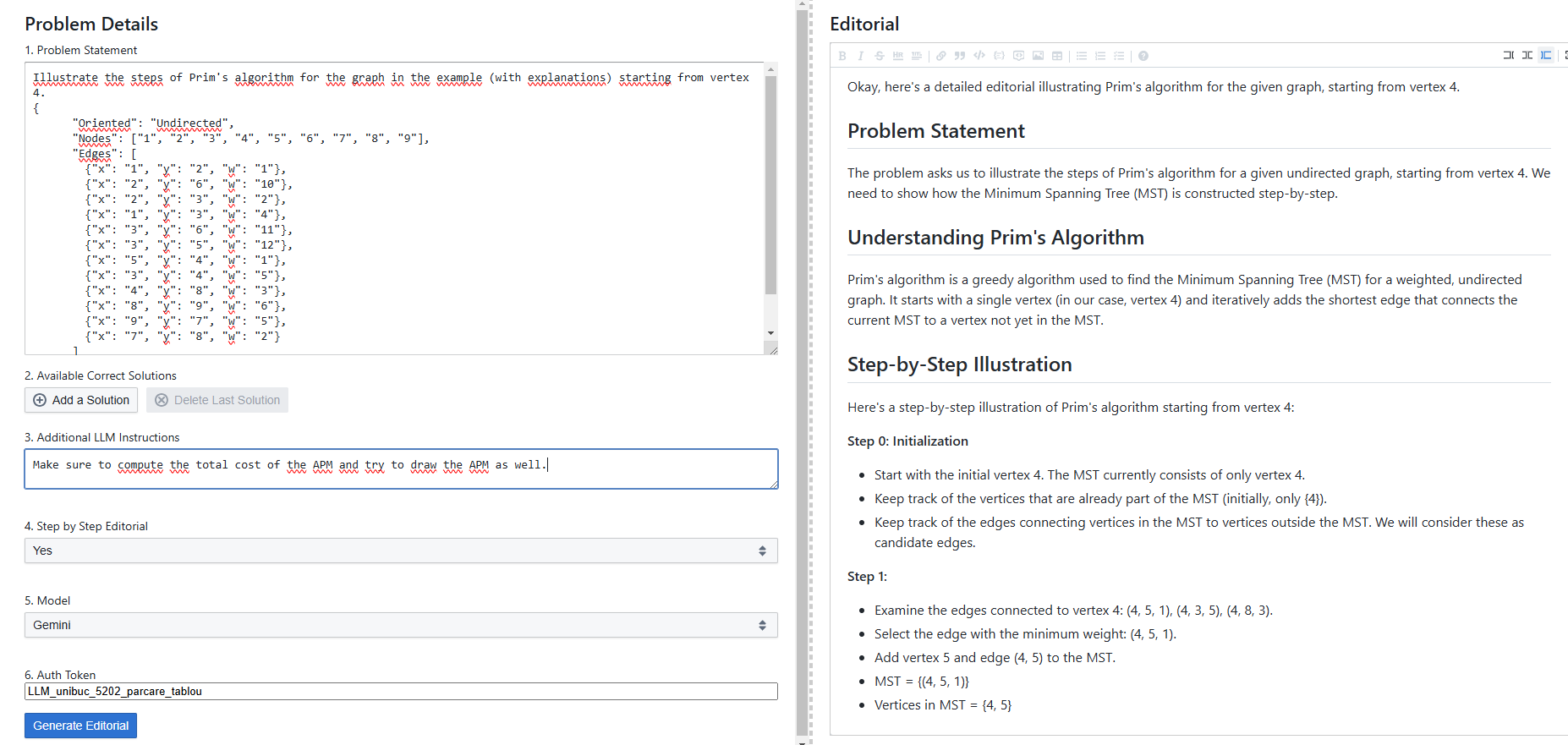}}
    \caption{Framework for generating solutions to a problem.}
    \label{fig:solution_framework}
\end{figure}

To showcase the practicality of our technique, we design and implement a web-based application, the design of which is presented in Figure \ref{fig:design}, and whose user interface can be seen in Figure \ref{fig:solution_framework}. For pricing reasons, our platform only supports \texttt{Gemini2.0 Flash} and \texttt{Mistral large}, which the users can freely pick when submitting a request. 

The application aims at being a tool tailored to professors and students work speeding up their editorial / grading scheme creation, respectively their learning experience, and is based on the observation that modern LLMs are sufficiently good and consistent for generating one-shot useful solutions, explanations and editorials for complex tasks. While it is currently more focused towards computer science, it can be easily modified to support a wide range of disciplines. 2 potential use-cases of the platform are:

\begin{itemize}
    \item \textbf{Editorial generation}: A history professor composed a written exam in German on the history of the Roman Empire, with the language choice made arbitrarily. After preparing a model solution outlining the expected answers, she uploaded both the exam questions and her solution to the platform.

    The system automatically generated a comprehensive editorial featuring step-by-step reasoning and detailed explanations for each exercise. Finding one explanation somewhat ambiguous, she added an instruction requesting a more in-depth analysis for that exercise and regenerated the editorial.
    \item \textbf{Solution generation}: A student received a geography assignment, but failed to come up with the right answers. He opens the framework, uploads the assignment's statement to the platform, provides its current ideas and explains what he approaches he tried, and the platform automatically generates a solution to the assignment.
\end{itemize}

The platform \cite{ref_moro} is not a commercial product, and is currently only intended to be used for research purposes, mainly due to a lack of authentication mechanisms. Exposing an unprotected application querying LLMs to the internet can incur unexpected costs due to an abusive usage of our API: attackers can use techniques such as \cite{prompt-injection} for bypassing our prompts and run arbitrary queries at our expense. To prevent this, we define an \textit{auth token}, which is stored in the browser's local storage (i.e. only needs to be entered once), which acts like a pre-shared secret between the researchers using the application and the backend hosting it.

As one can see in Figure \ref{fig:design}, the application is made out of two main components. The frontend is implemented in \textit{React} and heavily relies on the \textit{BlueprintJS} ui library. The backend is implemented in \textit{Python} and interacts with the frontend with the help of \textit{FastAPI} and with the \textit{Mistral} and \textit{Gemini} APIs using their specific \textit{Python} module.
For the purpose of this article, the most relevant part however is the prompt engineering \cite{white2023prompt}. This was done in an iterative process, by adding and tweaking existing prompts to make the LLMs behave in an expected, useful and consistent manner. Our prompts add on average $2000$ additional characters to the LLM inference requests we make to \textit{Gemini 2.0 Flash} and \textit{Mistral large}, and we offer users the possibility to insert additional prompts.

\section{Conclusion}

In this study, we have explored the progress of state-of-the-art Large Language Models (LLMs) in tackling complex, university-level algorithm exams. Our empirical evaluation places models such as \texttt{o3-mini} and \texttt{Gemini2.0 Flash} on-par with top-performing students, showcasing their robust reasoning skills and consistency across multi-step algorithmic challenges.

Our consistency analysis provides a metric demonstrating a strong correlation between LLM consistency and performance on algorithmic tasks, indicating that improved models exhibit greater consistency. Modern LLMs, excelling in both theoretical exercises and practical applications, can support educational environments by generating high-quality editorial content. The consistency and performance of these models make them valuable tools for instructors and students, offering detailed grading schemes and actionable feedback to enhance learning.

\section{Future work}

Our study reveals remaining challenges in LLMs, particularly in handling graph-based tasks and ensuring grading fairness and accuracy. Future research should address these limitations, exploring multimodal capabilities for seamless interpretation of visual and textual data to unlock LLMs' full potential for interactive and effective learning experiences.

Rapid advances in LLMs unlock exciting applications, particularly in automated essay scoring (AES), a field previously limited by models' grasp of complex tasks. While AES has evolved considerably since its introduction \cite{ref_article1,ref_article2,ref_proc1}, few studies examine LLM potential in advanced STEM grading. Future work should analyze LLM effectiveness in these contexts, overcoming challenges in processing graph-rich exams and handling low-resource languages. This requires improvements in OCR, specialized digital exam capture, or future LLMs with robust multimodal image interpretation. Addressing these challenges will be crucial for harnessing LLMs in advanced STEM assessment.

Another promising direction is the development of specialized feedback loops that utilize LLM-generated editorial content to provide detailed, step-by-step explanations for complex algorithmic problems. This approach could help students better understand advanced concepts and identify common pitfalls. Finally, conducting controlled studies to measure the impact of such editorial feedback on student learning outcomes in advanced algorithms courses will be essential for validating these techniques and establishing best practices for generative tutoring systems.

Another promising research direction emerging from this work is the use of collaborative LLMs to tackle tasks where individual models consistently struggle, as discussed in subsection~\ref{subsec:collaborative}.

\begin{credits}
\subsubsection{\ackname}
The authors gratefully acknowledge Softbinator Technologies for their support of this research.

\subsubsection{\discintname}
The authors have no competing interests to declare that are relevant to the content of this article.
\end{credits}

\appendix
\section{Advanced Algorithms Exam Statement}
\label{appendix:exam_statement}

\noindent
\textbf{Advanced Algorithms}\\
\begin{wrapfigure}[22]{l}{0.3\textwidth}
  \centering
  \includegraphics[width=0.28\textwidth]{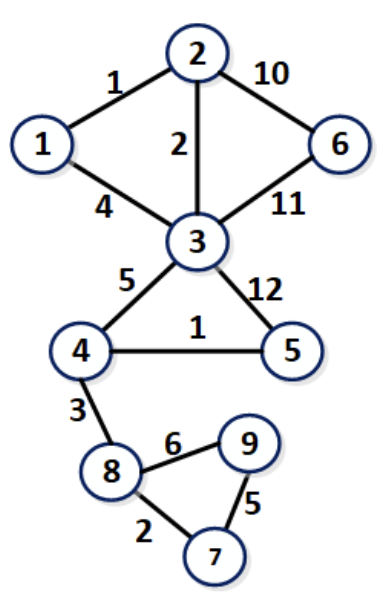}
  \caption{Graph representations for Exercises 1–6.}
  \label{fig:Graph1_6}
\end{wrapfigure}
For the graph given below, solve tasks 1--6 and justify your answers;
the neighbors of a vertex are considered in lexicographical order.

\begin{enumerate}

\item \textbf{(5p)} Specify a connected bipartite induced subgraph with the maximum number of nodes and a bipartition of it.

\item \textbf{(5p)} Illustrate (with explanations) how breadth-first search \(\mathrm{bfs}(4)\) works, showing how \(\mathrm{bfs}(4)\) can be used to calculate the distance from 4 to the other nodes. Draw the BFS tree as well.

\item \textbf{(7.5p)} Determine whether the graph admits an Eulerian path. If not, remove a minimum number of edges so that the resulting graph has an Eulerian path, describing the strategy you used to choose the removed edges. Specify an Eulerian path in the initial/obtained graph. State a necessary and sufficient condition for an undirected graph to have an Eulerian path.

\item \textbf{(7.5p)} Describe an efficient algorithm for determining the critical nodes of an undirected graph and exemplify (with explanations) the algorithm for the graph in the image.

\end{enumerate}

\begin{enumerate}
\setcounter{enumi}{4}

\item \textbf{(5p)} Describe the Floyd-Warshall algorithm for determining distances in a weighted undirected graph with \(n\) vertices, detailing the following scheme:
\begin{verbatim}
Initialize the distance matrix D with the cost matrix.
for j ← 1 to n do
   for i ← 1 to n do
      for k ← 1 to n do
         ...
\end{verbatim}
Write which values are modified in the matrix for the graph in the example at stages \(j=1\), \(j=2\), and \(j=3\) (with explanations).

\item \textbf{(5p)} Illustrate the steps of Prim's algorithm for the graph in the example (with explanations) starting from vertex 4.

\item \textbf{(5p)} Is the following algorithm for determining a minimum spanning tree of a connected weighted graph \(G = (V, E, w)\) correct? Justify (without appealing to the functioning of other algorithms in the justification; the used results must be demonstrated and it must be explained how they were used):

\begin{verbatim}
T = (V, E = <empty set>)  -- initially V contains all the 
vertices and contains no edge.
for i = 1 to |V|-1
   Choose the connected component C of T that contains vertex i.
   Choose an edge of minimum cost e with one end in C and 
   the other not in C and add e to T.
\end{verbatim}
\end{enumerate}

\begin{enumerate}
\setcounter{enumi}{7} 
    \item \textbf{(12.5p)} Illustrate the steps of the Ford-Fulkerson algorithm for this network starting from the indicated flow and choosing at each step an \(s\)-\(t\) \(f\)-unsaturated path of minimum length (the Edmonds-Karp algorithm). Indicate a minimum cut (\(s\)-\(t\) cut) in the network (the vertices in the bipartition, the direct arcs, the reverse arcs, and how it is determined by the algorithm) and determine the capacity of this cut. Justify the answers.
\end{enumerate}

\begin{enumerate}
\setcounter{enumi}{8}

\item \textbf{(15p)}
  \begin{enumerate}
    \item Show that a graph with \(n>2\) nodes that satisfies the condition \(d(x) \ge \frac{n}{2}\) for any node \(x\) is connected.
    \item Give an example of a non-Hamiltonian graph in which there are two distinct non-adjacent nodes with the sum of degrees \(\ge n\).
    \item Show that if a graph \(G\) with \(n \ge 2\) nodes has \(m \ge \binom{n-1}{2} + 2\) edges, then \(G\) is Hamiltonian.
  \end{enumerate}

\item \textbf{(7.5p)} Briefly describe the algorithm for determining the maximum length of a common subsequence of two words. Illustrate the algorithm for the words \emph{cerceta} and \emph{retea} by writing the matrix with the values of the subproblems and explaining how they were calculated.

\item \textbf{(15p)} A team of explorers has discovered an old map of an underground mine renowned for a rare and valuable crystal. The mine is composed of a series of chambers interconnected by unidirectional tunnels. For our experienced explorers, the tunnels can be traversed without any effort. However, some of the chambers have collapsed, and to cross them, they need to use dynamite. The team's goal is to get from the entrance chamber to the chamber containing the rare crystal using as little dynamite as possible. Write an algorithm of optimal complexity that determines if there is a path for the explorers and, if there is, determines the path. 
(7.5p for explaining the solution and 7.5p for complexity analysis)
\end{enumerate}

\bibliographystyle{splncs04}
\bibliography{references}
\end{document}